# CacheFormer: High Attention-Based Segment Caching

Sushant Singh, Ausif Mahmood

**Abstract**

Efficiently handling long contexts in transformer-based language models with low perplexity is an active area of research. Numerous recent approaches like Linformer, Longformer, Performer, and Structured State Space Models (SSMs)., have not fully resolved this problem. All these models strive to reduce the quadratic time complexity of the attention mechanism while minimizing the loss in quality due to the effective compression of the long context. Inspired by the cache and virtual memory principle in computers, where in case of a cache miss, not only the needed data is retrieved from the memory, but the adjacent data is also obtained, we apply this concept to handling long contexts by dividing it into small segments. In our design, we retrieve the nearby segments in an uncompressed form when high segment-level attention occurs at the compressed level. Our enhancements for handling long context include aggregating four attention mechanisms consisting of short sliding window attention, long compressed segmented attention, dynamically retrieving top k high attention uncompressed segments, and overlapping segments in long segment attention to avoid segment fragmentation. These enhancements result in an architecture that outperforms existing SOTA architectures with an average perplexity improvement of 8.5% over similar model sizes.

## Introduction

Deep Convolutional Neural Networks (CNNs) were fundamental in revolutionizing the field of computer vision. Similarly, the pioneering induction of the Transformer (Vaswani et al., 2017) architecture in Natural Language Processing (Singh and Mahmood, 2021) has resulted in the AI revolution with Large Language Models (LLMs) such as ChatGPT (J. Achiam et al., 2023), Bard (G. Team et al., 2023), Llama (H. Touvron et al., 2023) among others have yielded impressive performances. The Transformer uses a simple similarity computation in the form of an inner product on the learnt positional encoded embeddings of a sequence of $n$ input words. If the matrix $Q$ and $K$ contain rows representing embedding of each word $(1 \times d)$, then $A = softmax(QK^T)$ referred to as the "attention", contains the dot product similarity of each input word with every other word in the input sequence. If there are $n$ words being input, referred to as the context, then $Q, K \in \mathbb{R}^{n \times d}$, and $A \in \mathbb{R}^{n \times n}$.

Like parallel feature maps in a CNN, each layer in the Transformer divides the attention calculation into parallel heads. The output from a Transformer layer has the same dimensionality as input and is obtained by a simple matrix computation of $(A \times V) \in \mathbb{R}^{d \times n}$ where $V \in \mathbb{R}^{n \times d}$ is similar to $K$ and contains rows of learnt position encoded embeddings of input words. For language models, where text generation is carried out based on a given context, the attention matrix is masked in a triangular fashion so that future tokens are not visible in the training process. Multiple layers of Transformer blocks are used before feeding the result of the last layer to a classification head. Because attention computation in each head is $O(n^2)$, for long contexts, this becomes a computational bottleneck. Many approaches have been proposed in the past years to reduce the quadratic time complexity of attention to either linear or sub quadratic complexity. Some of the notable works include Transformer-XL (Dai Z et al., 2019), Linformer (Wang et al., 2020), Longformer (Beltagy et al., 2020), Reformer (Kitaev et al., 2020), Performer (Choromanski et al., 2021), Perceiver-AR (Hawthorne et al., 2022), LaMemo (H. Ji et al., 2022), ∞-former (Martins et al., 2022) among others. We provide a brief background in the above-mentioned approaches used in reducing the attention complexity. Then we elaborate on the Long-Short Transformer (C. Zhu et al., 2021) that we further enhance in this work.

Although Long-Short Transformer performs efficient compression on the input sequence, but this compression results in segment fragmentation that leads to these key shortcomings.

- It reduces the dimensionality of the input sequence resulting in a loss of context.
- The original sequence is divided into smaller, potentially non-overlapping segments. Since information is isolated within segments, it disrupts the natural flow and continuity of information.

These two factors make it more challenging for the language model to capture overall context and relationships between distant elements in the sequence. Our architecture in CacheFormer alleviates some of the key constraints that can be summarized as:

- We developed an innovative uncompressed attention mechanism where the highly attentive segments are dynamically cached in an uncompressed form.
- Improved segment fragmentation of the projection mechanism followed in long attention. We achieved it by adding projections of segments that have an $s/2$ overlap where $s$ is the segment size, with the existing segment-based projection mechanism as illustrated in Figure 3.

- We combine the existing short and long attention with our above two enhancements in an effective manner. This results in an architecture that can efficiently handle long attentions without causing much loss of attention information.

## Background and Related Work

An important earlier work in handling long contexts was presented by (Dai Z et al., 2019). The authors divided the context into segments and used segment level recurrence and a corresponding positional encoding to allow it to handle longer contexts. It achieved impressive results on the perplexity and BPC at that time. (Wang et al., 2020) accomplished $O(n)$ complexity through linear self-attention. The authors demonstrate that the attention is typically low rank, and thus can be approximated by a low rank matrix. Here, the Q and V matrices $\in \mathbb{R}^{n \times d}$ are projected to lower dimension matrices $\in \mathbb{R}^{k \times d}$ where $k < n$. Thus attention $A = QK^T \in \mathbb{R}^{n \times k}$. The output $(A \times V) \in \mathbb{R}^{n \times d}$, i.e., same as the original transformer. Since $k$ is fixed, the attention complexity is $O(n)$.

Although (Wang et al., 2020) reduced the attention complexity significantly, especially if $k << n$, note that, it cannot be effectively used in autoregressive training and generation, as the projection of Q compresses the information along the context, making the masking of attention for future tokens invalid. However, for classification problems where masking of attention is not needed, their architecture is effective in reducing complexity.

Another approach introduced by (Beltagy et al., 2020) used sparse attention patterns instead of the full dense attention. The authors proposed sliding window attention, where tokens attended only to the nearby past, a dilated sliding window, and a mix of global and sliding window attention where some tokens attend to all tokens while others only attend to nearby tokens. For autoregressive modeling (Beltagy et al., 2020) used dilated sliding window attention. Another notable work in reducing the attention complexity was performed by (Kitaev et al., 2020). The authors key idea was to use locality sensitive hashing which reduces the attention complexity to $O(n \log n)$. Note that because of the hashing process, the architecture is not suited for autoregressive modeling.

A different approach to reduce the attention complexity was taken by (Choromanski et al., 2021) where the attention is decomposed as a product of non-linear functions of original query and key matrices referred to as random features. This allows the attention to be encoded more efficiently via the transformer query and key matrices. Further efficient handling of long contexts accomplished by (Hawthorne et al., 2022) divided the input sequence into smaller key/value and query components. These components underwent cross attention in the first layer with a latent $\in \mathbb{R}^{l \times d}$ where $l$ is the size chosen in splitting the input sequence into the query part. The remaining layers operate on the $l \times d$ size instead of the usual $n \times d$ size as in a standard transformer. Although this cross attention on the partitioned input sequence results in efficient handling of long sequences, because of the reduced query size, the equivalent effect is more like a sliding window attention. More recently, a different approach to handling long contexts was proposed via structure state space models. The work by (A. Gu et al., 2022) proposes the Structured State Space Sequence model (S4) based on a new parameterization that can be computed much more efficiently. A variation of the state space approach proposed by (X. Ma et al., 2023) uses single-head gated attention mechanism equipped with exponential moving average to incorporate inductive bias of position-aware local dependencies into the position-agnostic attention mechanism. They also present its variation with linear time complexity for handling long sequences. Further progression on the state space models yielded better results (D. Y. Fu et al., 2023), (Gu and Dao., 2023) who achieved a very low perplexity score. Most recently (Maximilian et al., 2024) introduced exponential gating and parallelization in LSTMs to achieve extended memory. Some of the model sizes consisted of several billion parameters. We outperform the smaller version of these models with similar size as ours on the perplexity metric as shown in Table 2.

An interesting concept in handling long sequences was presented by (C. Zhu et al., 2021). Here a sliding window approach is used in handling near term attention, while a set of compressed segments for the entire past context is used as long-term attention. Both short and long attention are combined in the overall attention. The slight drawback of the approach is that the longer context is effectively used in compressed form and thus may lose some key contextual information in being able to generate the output in an autoregressive environment. We address this problem by further augmenting the long-short attention by using uncompressed highly attentive segments. Since long short attention divides the context into equal size segments before projecting each segment to a smaller size, there is potential for a loss in information due to segment fragmentation. We also improve this aspect by using overlapping segments and augment this to the existing long-short model. Thus, our enhanced long-short architecture involves four components in the overall attention, a sliding window attention, long attention based on compressed segments, long attention based on overlapping segments, and uncompressed segmented attention for few high attentive segments beyond the sliding window part. We describe the details of our design in the section 3. For completeness, we summarize the composition of a Transformer, followed by the ideas of long-short Transformer, that we build upon in our work.

## Canonical Transformer

In normal multi-headed attention, if $Q, K, V \in \mathbb{R}^{n \times d}$ are the query, key and value transformations of the input embeddings with sequence length of *n* and embedding dimension of *d*, then the scaled dot-product attention in the *i*-th Head $H_i \in \mathbb{R}^{n \times d_k}$ is given as:

$$H_i = Attention(QW_i^Q, KW_i^K, VW_i^V) = Softmax\left[\frac{QW_i^Q(KW_i^K)^T}{\sqrt{d_k}}\right] VW_i^V = A_i VW_i^V \quad (1)$$

where $d_k = d/h$ is the dimension of each head. The output in each transformer layer is obtained by catenation of the output of all heads and transformed further via this projection matrix.

$W^o \in \mathbb{R}^{d \times d}$ as $Layer_j =$
$$Concat(H_0, H_1, \cdots H_{h-1}) W^o \quad (2)$$

After feeding the embedding of a sequence of one hot encoded word, *x* (with position encoding *PE* added) through *p* transformer layers, a classification layer is used at the output of the last layer to decide the output produced by the transformer. For autoregressive text generation, the classification layer's final output is equal to the size of the dictionary of unique words in the corpus.

$$out = classifier[layer_{p-1}(layer_{p-2}(\ldots layer_0 (embedding(x) + PE(x))))] \quad (3)$$

## Long Short Transformer

(C. Zhu et al., 2021) aggregated the local attention around a smaller window (sliding window), with a projection of the full sequence attention to a smaller size, so that we can efficiently handle long sequences without the quadratic attention complexity. For short attention, the approach here is to use a segment level sliding window attention, where the input sequence is divided into disjoint segments with length *w* (e.g., *w*=128 and sequence length is 1024). For non-autoregressive applications, all tokens within a segment attend to all tokens within its home segment, as well as w/2 consecutive tokens on the left and right side of its home segment (zero-padding when necessary), resulting in an attention span over a total of 2*w* key-value pairs. This is depicted in Figure 1.

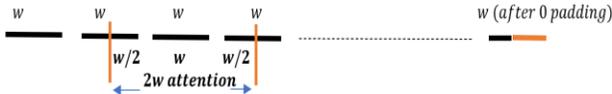

Figure 1. Segment-based Sliding Window Attention

For each query $Q_t$ at the position *t* within the $i^{th}$ head, the 2*w* key-value pairs within its window are: $\widetilde{K_t}, \widetilde{V_t} \in \mathbb{R}^{2w \times d}$.

The short attention $\bar{A}_{s_i} \in \mathbb{R}^{2w \times d_k}$ is then given by the following equation:

$$\bar{A}_{s_i} = softmax\left[\frac{QW_i^Q \widetilde{K}_i^T}{\sqrt{d_k}}\right] \quad (4)$$

Execution wise the segment-level sliding window attention (referred to as short attention) is more time efficient than the per-token sliding window attention where each token attends to itself and *w* tokens to its left and right, and its memory consumption scales linearly with sequence length. For auto-regressive applications, the future tokens in the current segment are masked, and only the previous segment is used.

The compression is performed on the feature dimension initially through a projection with dimensions, $p \rightarrow (d_k \times r)$ where $d_k$ is the embedding dimensionality and *r* is the target length. The dynamic projection matrix $'P_i'$ is computed by multiplying *p* with *n* length Key (K) $\rightarrow (n \times d_k)$ where $(r \ll n)$ i.e.($K \times p$). This product results to $'P_i'$ with dimensions $(n \times r)$. The transpose of this projection matrix $P_i^T$ is applied to the Key vector $\rightarrow (p^T \times K)$. This product results to a modified Key $(\overline{K})$, with dimensionality $\rightarrow (r \times d_k)$ thereby compressing its sequence length. Similar compression is performed for the Value vector as well. This is a standard dimensionality reduction technique illustrated in Figure 2 and is used in popular models like Performer and Transformer LS.

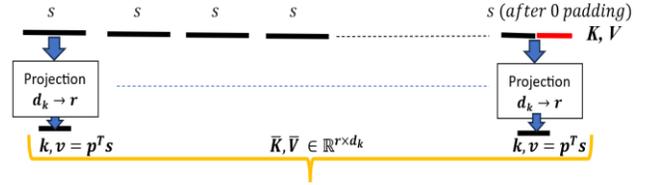

Figure 2. Segmented Long Attention with Compressed Segments

For long attention, the key and value transformations for the input sequence are first divided into segments of fixed size *s*, and then projected to a smaller dimension *r*, where the projection $P_{l_i} \in \mathbb{R}^{n \times r}$.

Mathematically, the long attention $\overline{A_{l_i}}$ (in each head *i*) as followed by the long-short Transformer can be described as
$P_{l_i} = Softmax(KW_i^P), \overline{K}_{l_i} = P_{l_i}^T KW_i^K, \overline{V}_{l_i}$
$$= P_{l_i}^T VW_i^V \quad (5)$$

$$\bar{A}_{l_i} = softmax\left[\frac{QW_i^Q \overline{K}_{l_i}^T}{\sqrt{d_k}}\right] \quad (6)$$

The output of in the $i^{th}$ head is:
$$\overline{H}_i = \bar{A}_{l_i}\left(P_{l_i}^T VW_i^V\right) \quad (7)$$

Note that the long attention is effectively done on a compressed form of *K* and *V*, as the projection causes the input sequence of size *n* to be compressed to size *r*. This results in full attention to now be replaced with the implicit product of two low-rank matrices $\overline{P_{l_i}^T} \in \mathbb{R}^{r \times n}$ and $QW_i^Q \in \mathbb{R}^{n \times d}$, and thus the computational complexity is of long attention is reduced from $O(n^2)$ to $O(rn)$.

Long-Short Transformer integrates the short and long attentions into a single attention. While the short attention can attend to most recent input, the long attention is in compressed form. Further, the long attention is based on segmentation of the input sequence that may suffer from segment fragmentation as the information in each segment is compressed via the projection mechanism.

## Enhanced Long-Short Transformer

The long-term attention in the existing Long-Short Transformer is done at a compressed level (projection to *r* causes an effective compression of the input context). Therefore, one of our enhancements is to augment the long attention with an attention that is based on a subset of highly attentive uncompressed segments.

### Enhanced Long Attention with Segment Caching

The subset of segments that are selected for attention at the uncompressed level is completely dynamic and obtained by the vector magnitude of the compressed segment-wise attention. In simple words, we examine the segment-wise long attention $\bar{A}_{l_i}$ as given by Equation 6. Since $\bar{A}_{l_i} \in \mathbb{R}^{n \times r}$, and if there are $n_s$ segments, then each row in $\bar{A}_{l_i}$ contains a set of row vectors of size $r/n_s$, as denoted by segmented attention $\overline{A_{seg_i}}$ in Equation 8. Magnitude of each vector $\overrightarrow{a_{i,j}} \in \mathbb{R}^{1 \times r/n_s}$ in Equation 8, indicates the attention of word $i$ to the $j^{th}$ segment in the long attention.

For execution efficiency, we average the segment attention vectors in $p$ consecutive rows resulting in a segment attention matrix $A_{sega_{vgi}} \in \mathbb{R}^{m \times n_s}$ where $m = n/p$. Then we choose top $k$ segments by magnitude of each vector in each row of the segment attention matrix $A_{segavg_i}$. Note that each entry in the segment attention matrix, $Aeg_{sa_{vgi}}[i,j]$, indicates the segment number that has high attention to the sequence of $p$ words positioned from $[((i - 1) \times p)$ to $(i \times p)]$ in the input context. Rather than using these attentive segments in compressed form, we extract them from the segmented $K$ and $V$ matrices before doing any compression on these.

$$\overline{A_{seg_i}} = \begin{bmatrix} \overrightarrow{a_{1,1}} & \overrightarrow{a_{1,2}} & .. & .. & \overrightarrow{a_{1,n_s}} \\ :: & :: & .. & .. & :: \\ \overrightarrow{a_{n,1}} & \overrightarrow{a_{n,2}} & .. & .. & \overrightarrow{a_{n,n_s}} \end{bmatrix} \quad (8)$$

$$A_{segavg_i} = \begin{bmatrix} \overrightarrow{topk((\sum_{t=1}^{p} \overline{A_{s_t}}[t,:])/p)} \\ \overrightarrow{topk((\sum_{t=p+1}^{2p} \overline{A_{s_t}}[t,:])/p)} \\ .. \\ .. \\ \overrightarrow{topk(((\sum_{t=n-p}^{n} \overline{A_{s_t}}[t,:])/p)} \end{bmatrix} \quad (9)$$

Similar to how in cache memory design (in computer architecture), in case of a cache miss, we not only retrieve the needed data from the RAM, but also bring a few consecutive following words, as there is high probability that these may be needed in the near future. In case of segments that we determine most attentive (by the top *k* order), we also retrieve *u* consecutive segments. To clarify our approach, if the sequence length is $n = 1024$, and long attention segment size $= 16$, then there will be 64 segments in the uncompressed $K$ and $V$ matrices. If the projection size $r = 256$ (ratio of 1024/256=4), then each segment of size 16 will be compressed to size of 4, resulting in long attention matrix $\overline{A_{l_i}}$ of size 1024x(64x4) i.e., 1024x256. If we choose to average $p=32$ consecutive rows in $\overline{A_{l_i}}$, and take the magnitude of each of the 1x4 vectors in each row (corresponding to the 64 segments), then the segment attention matrix $A_{sega_{vgi}}$ will be 32x64. Taking the index of top *k* entries in each row of $A_{segavg_i}$ will give us the index of most attentive *k* segments to the corresponding set of 32 words in the input sequence. Assembling these top *k* attentive segments, and one segment before and one segment after the attentive segment (if $u=3$), will result in 15 segments per row. If $k=5$ is chosen in *top-k* and $u=3$ which indicates using of $u$-1 many nearby segments for each attentive segment. Thus, the cache $K, V$ matrices $K_c, V_c \in \mathbb{R}^{(n/p) \times (k \times u)}$ (e.g., 32x(15x16) = 32x240 in this case) contain the most attentive 15 segments in uncompressed form. Note that we stack the $K_c$ '*p*' times to match the dimensionality with $Q$. From the most attentive $k \times u$ segments in $K_c$, we can obtain the cache attention $\bar{A}_{c_i} \in \mathbb{R}^{n \times (k \times u)}$ as:

$$\bar{A}_{c_i} = softmax(QW_i^Q \begin{bmatrix} K_c \\ K_c \\ :: \\ K_c \end{bmatrix}^T )/\sqrt{d_k} \quad (10)$$

### Enhanced Long Attention with Overlapping Segments

In addition to the original long attention in the Long-Short Transformer that uses the projections on each segment, we augment the existing long attention by using overlapping segments (with 50% overlap in augmented long attention) as shown in Figure 3. The motivation behind the overlap is to reduce the effect of segment fragmentation in long attention. Zero padding in the beginning segment is added to ensure the same dimensionality for the overlapped long segment attention. The overlapped long segment attention $\bar{A}_{o_i} \in \mathbb{R}^{n \times r}$ similar to Equation 5 is given below.

$$P_{o_i} = Softmax(KW_i^{Po}), \bar{K}_{o_i} = P_{o_i}^T KW_i^K$$
$$\bar{V}_{o_i} = P_{o_i}^T VW_i^V \tag{11}$$

$$\bar{A}_{o_i} = Softmax\left[\frac{QW_i^Q \bar{K}_{o_i}^T}{\sqrt{d_k}}\right] \tag{12}$$

$$\bar{H}_{o_i} = \bar{A}_{o_i} \left(P_{o_i}^T VW_i^V\right) \tag{13}$$

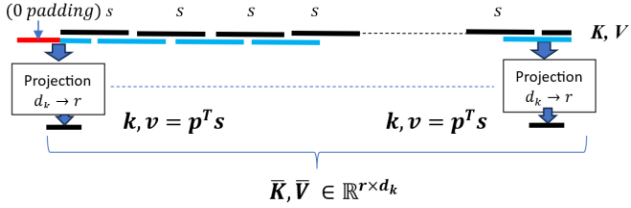

**Figure 3.** Overlapping Segmented Long Attention with Compressed Segments

## Aggregated Long-Short Attention

The final attention in our enhanced architecture is obtained by aggregating the four attentions as discussed earlier:
1) The short attention, $\bar{A}_{s_i} \in \mathbb{R}^{n \times 2w}$ that uses segment-wise sliding window in Transformer Long Short.
2) The segment based compressed long attention, $\bar{A}_{l_i} \in \mathbb{R}^{n \times r}$ as proposed in Transformer Long Short.
3) Our cache attention based on dynamic retrieving of uncompressed high attention segments, $\bar{A}_{c_i} \in \mathbb{R}^{n \times (k \times u \times s)}$
4) Our overlapping segment-based compressed attention, $\bar{A}_{o_i} \in \mathbb{R}^{n \times r}$.

We add the two similar sized long and overlapping attentions, $\bar{A}_{l_i}$ and $\bar{A}_{o_i}$, and ∥ indicates the catenation of different sized attentions, $\bar{A}_{s_i}$ and $\bar{A}_{c_i}$. Thus, the final enhanced attention $A_{e_i} \in \mathbb{R}^{n \times f}$ is expressed as:

$$A_{e_i} = [\bar{A}_{s_i} \| (\bar{A}_{l_i} + \bar{A}_{o_i}) \| \bar{A}_{c_i}] \tag{14}$$

where $f = 2w + r + (k \times u \times s)$.
➤ $w$ is the window size in short i.e., sliding window attention.
➤ $r$ is the compressed projection target size in the long attention.
➤ $top\ k$ factor is for retrieving top $k$ attentive segments.
➤ $u$-1 is the number of neighboring segments to retrieve for cache attention
➤ $s$ is the segment size in long attention.

For example, in $top\ k = 5$; $u = 3$; segment size in short attention, $w = 128$; segment size in long attention, $s = 16$; compression target length, $r = 256$. Hence, for an input sequence length of 2048, the size of our combined attention matrix is 2048x752.

The time complexity of the different components in our Enhanced Long-Short attention is as follows:
- For the short attention $\bar{A}_{s_i} \rightarrow O(w \times n)$, where $w$ is the sliding window size.
- For both long and overlapping long attentions i.e., $\bar{A}_{l_i}, \bar{A}_{o_i}, \rightarrow O(p \times m \times n)$, where $p$ is the compressed output size from each of the $m$ long segments
- For cached attention $\bar{A}_{c_i}, \rightarrow O(k \times u \times s \times n)$, where $k \times u$ is the number of the top attentive segments, and $s$ is the long attention segment size.

Since the dominant term in the above four components is the long attention, the overall time complexity of our enhanced attention is $O(p \times m \times n)$. Effectively, this is very close to the sliding window attention.

To further elaborate on our attention computation in Equation 14, note that the dimensionality of short sliding attention $\bar{A}_{s_i}$ in the LS Transformer is $(n \times 2w)$ and its compressed long attention's, $\bar{A}_{l_i}$ dimensionality is $(n \times r)$. During our caching mechanism, we augment attentions $\bar{A}_{o_i}$ and $\bar{A}_{c_i}$ with dimensionalities $(n \times r)$ and $n \times (k \times u \times s)$ respectively to the Long Short attention. Since $\bar{A}_{l_i}$ and $\bar{A}_{o_i}$ deal with sequence lengths compressed to similar dimensions, they have similar shapes. Therefore, we can sum up the two attention matrices along the similar dimensions to conserve size and overall attention complexity. Whereas our caching attention $\bar{A}_{c_i}$ and $\bar{A}_{s_i}$ have different shapes, hence they cannot be summed up and concatenation is the only choice.

## Results

We use the long-short transformer (C. Zhu et al., 2021) as the baseline architecture. Instead of focusing on the absolute best results for perplexity and BPC, which often are achieved through extremely refined training schedules and large model sizes, we focus on the improvements over the baseline. Therefore, the results we show are more accurate reflection of the architectural improvements of our design. The baseline architecture is also programmed by us, and the enhancements we propose are programmed in the same implementation and can be selectively turned on or off to see the contribution of each enhancement. We also use similar training schedules for the different architectures being compared. Table 1 shows the perplexity results for wikitext-103 dataset. It uses sequence length of 1024, short attention segment size of 128, long attention segment size of 16, compression of the long sequence by a factor of 4, i.e., $r$=256, and different values of $k$ in top k cache attention, and neighboring segments retrieval u of 1 or 3 (which indicates the segment before the attentive segment, and the one after it is also retrieved.

Note that our enhanced architecture does not cause any increase in the number of model parameters over the baseline long short Transformer. The models used for results in Table 1 have 12 layers, 12 heads, and an embedding size of 768

(for all architectural variations). For a sequence length of 1024 (which is same as used in GPT-2), using 7 segments ($k=7$, $u=1$) yielded considerable improvement in perplexity. Increasing k beyond 7 did not seem to considerably reduce perplexity further. Since we have two major enhancements of cache attention and overlapping segment-based attention over the baseline, Table 2 shows an ablation study of the effects of each architectural improvement.

Figure 4 depicts the 64 attention vectors for each segment (from compressed long attention, after averaging $p=256$ rows) corresponding to the 64 segments during the beginning of training. The highest top $k$ magnitude vectors then determine the segment to use in uncompressed form for our cache attention. Table 3 shows the BPC results on the enwik-8 benchmark. The 23 million model uses 8 layers, 8 heads and embedding size of 512. The 34.88 million models used 12 layers. It is interesting to note that the relative improvement in BPC by our enhanced architecture is less pronounced as compared to the perplexity improvements. This could be attributed to the fact that majority of improvements are attributed to cache attention which uses a few highly attentive uncompressed segments in long attention.

| Model | Model Size | Perplexity |
|---|---|---|
| Long-Short Baseline | 122.52 million | 23.74 |
| Enhanced Long-Short ($k=3$, $u=1$) | 122.52 million | 23.31 |
| Enhanced Long-Short ($k=5$, $u=1$) | 122.52 million | 22.75 |
| Enhanced Long-Short ($k=7$, $u=1$) | 122.52 million | 21.32 |
| Enhanced Long-Short ($k=5$, $u=3$) | 122.52 million | 21.26 |

**Table 1.** Perplexity results Comparing the Baseline and our Enhanced Architecture

While this benefits the perplexity which is a measure of the model's prediction capability, but BPC not as much, as BPC is more of a compression efficiency measure of the model.

| Architecture | Model Size (Millions) | Perplexity |
|---|---|---|
| Long-Short (Baseline-Ours) | 122.52 | 23.74 |
| Transformer-XL (Standard) | 151 | 24 |
| ∞-former | 160 | 24.22 |
| LaMemo | 151 | 23.77 |
| H3 (Hungry Hungry Hippos) | 125 | 23.7 |
| Llama | 125 | 23.16 |
| Mamba | 125 | 22.49 |
| xLSTM[7:1] | 125 | 21.47 |
| Enhanced Long Short with overlapping segments only | 122.52 | 23.47 |
| Enhanced Long Short with cache attention only ($k=7$, $u=1$) | 122.52 | 21.67 |
| Enhanced Long Short with overlapping segments and cache attention ($k=7$, $u=1$) | 122.52 | **21.32** |

**Table 2.** Ablation Study of Architectural Enhancements

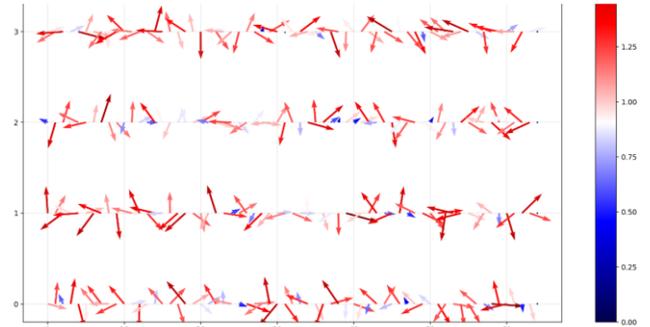

**Figure 4.** Attention Vectors from Compressed Long Attention

| Model | Model Size | BPC |
|---|---|---|
| Long-Short Baseline | 23 million | 1.192 |
| Enhanced Long-Short ($k=7$, $u=1$) | 23 million | 1.188 |
| Long-Short Baseline | 34.88 million | 1.173 |
| Enhanced Long-Short ($k=7$, $u=1$) | 34.88 million | 1.167 |

**Table 3.** Comparison of BPC on the enwik-8 Benchmark

## Discussion

Since the uncompressed segments to be used in our cache attention design are dynamically decided based on the input sequence, the execution time increases as more segments (i.e., higher k) are used. When we use, sequence length of 1024, compression $r = 256$, $k = 7$, $u = 1$, short attention segment size of 128, then the size of aggregated attention (short, long, cache, overlapping) is 1024x624. Since our cache attention mechanism as explained in section 3.1 is completely dynamic, and uses the most attentive segments in uncompressed form, we average the attention vectors over $p$ rows (to improve efficiency of execution) as given by Equation 9. If we use a sequence length of 1024, and average over 256 rows, then the segments determined by our cache attention mechanism part way through the training of the model appears as shown in Table 4. Note that to implement the autoregressive behavior, the input sequence cannot attend to a future segment. Our implementation guarantees that the input sequence can only attend to a previous segment. For example, when attending to words 768-1023 in the input sequence, the maximum segment that the cache attention can use is 47 (if the long segment size is 16, then there are 64 segments in the 1024 size sequence).

One of the important recent papers in handling long contexts has indicated that current language models do not robustly make use of information in long input contexts (N. F. Liu et al., 2023). They studied different models and concluded that "performance is often highest when relevant information occurs at the beginning or at the end of the input context, and significantly degrades when models must access relevant information in the middle of long contexts."

| Input Sequence | Top k Attentive Segments ($k=7, u=1$) | Comments |
|---|---|---|
| 0 – 255 words | [ -1, -1, -1, -1, -1, -1, -1] | No cache segments are used to prevent future token leakage |
| 256-511 words | [ 7, 8, 11, 12, 13, 14, **15**] | Maximum segment allowed = **15** |
| 512-767 words | [ 7, 8, 27, 28, 29, 30, **31**] | Maximum segment allowed = **31** |
| 768-1023 words | [ 8, 29, 32, 35, 37, 44, **47**] | Maximum segment allowed = **47** |

**Table 4.** Most Attentive Segments Used by our Cache Attention Part way in Training.

Note that our cache attention model addresses this aspect nicely in the sense it uses attentive segments dynamically regardless they are needed in the beginning or the middle of input context. For example, the last row in Table 4 indicates the highest attentive segments that are used. Segments 32, 35, 37 are relatively in the middle of the input context. When we determine the most attentive segment to use in our cache attention, if the neighboring segment parameter count $u>1$, then as we look at the segment index of the next or previous index, a duplicate may occur as the next segment may already be one of the high attentive segments. Similarly, if the high attentive segments belong to a future segment, we replace them by one of the allowed segments. Since information segmentation should not occur, the segment we select to be added is the one that is contiguous to an existing high attention segment.

## Conclusions

Handling long contexts in an efficient manner without loss of performance is an important area of research in language models. Although many approaches have been recently proposed to address this problem, we present a new innovative solution that is motivated by the cache and virtual memory concepts in computer architecture. In such designs, if there is a cache or page miss, the needed data is retrieved from the disk or RAM. We handle long contexts by diving them into small segments. By the magnitude of the compressed attention vectors, we determine the most attentive segments, and then use these in uncompressed form. Similar to the cache memory design, we also use consecutive segments near to the high attention segments to improve the language model predictive performance. Our results on the perplexity indicate significant improvement over the baseline architecture that uses short and long compressed attention. For the BPC, the cache attention mechanism does not show remarkable improvement on the baseline. We conjecture that the BPC that favors compression capability is not benefited by the relevant segment usage that our model provides which is helpful in model prediction capability. Another advantage of our approach is that the use of high attention segments is dynamic and depends on the input sequence. Thus, if the model needs to use information in the middle or anywhere in the input context, it is provided in uncompressed form via the high attention determination on the compressed segments.

## Limitations

The only shortcoming of our approach we feel is that the dynamic segment attention is relatively slow during training. We partially overcome this by initially pretraining the model without dynamic attention, and then fine tune it on our dynamic cached attention. Our future work involves in applying the cache attention to reduce the model complexity of large language models. Further we are in the process to create a hierarchical cache design such that very long contexts can be efficiently handled.

Further, our model sizes and datasets were constrained by computational resources available to us. We used GPU RTX 4090 and therefore could not use larger datasets such as PG-19 and run larger models with larger embedding size, layers, and heads.

## References

Vaswani, A.; Shazeer, N.; Parmar, N.; Uszkoreit, J; Jones, L.; Gomez, A. N.; Kaiser, L.; and Polosukhin, I. 2017. Attention is all you need. Proceedings of Neural Information Processing Systems.

Achiam, J. et al. 2023. OpenAI. GPT-4 Technical Report.

Team, Gemini, et al. 2023. Gemini: a family of highly capable multimodal models." *arXiv preprint arXiv:2312.11805*.

Touvron, H.; Lavril, T.; Izacard, G.; Martinet, X.; Lachaux, M.; Lacroix, T.; Rozière, B.; Goyal, N.; Hambro, E.; Azhar, F.; Rodriguez, A.; Joulin, A.; Grave, E.; and Lample, G.

2023. LLaMA: Open and Efficient Foundation Language Models. *ArXiv, abs/2302.13971*.

Touvron, H.; Martin, L.; Stone, K.R.; Albert, P.; Almahairi, A.; Babaei, Y.; Bashlykov, N.; Batra, S.; Bhargava, P.; Bhosale, S.; Bikel, D.M.; Blecher, L.; Ferrer, C.C.; Chen, M.; Cucurull, G.; Esiobu, D.; Fernandes, J.; Fu, J.; Fu, W.; Fuller, B.; Gao, C.; Goswami, V.; Goyal, N.; Hartshorn, A.S.; Hosseini, S.; Hou, R.; Inan, H.; Kardas, M.; Kerkez, V.; Khabsa, M.; Kloumann, I.M.; Korenev, A.V.; Koura, P.S.; Lachaux, M.; Lavril, T.; Lee, J.; Liskovich, D.; Lu, Y.; Mao, Y.; Martinet, X.; Mihaylov, T.; Mishra, P.; Molybog, I.; Nie, Y.; Poulton, A.; Reizenstein, J.; Rungta, R.; Saladi, K.; Schelten, A.; Silva, R.; Smith, E.M.; Subramanian, R.; Tan, X.; Tang, B.; Taylor, R.; Williams, A.; Kuan, J.X.; Xu, P.; Yan, Z.; Zarov, I.; Zhang, Y.; Fan, A.; Kambadur, M.; Narang, S.; Rodriguez, A.; Stojnic, R.; Edunov, S.; and Scialom, T. 2023. Llama 2: Open Foundation and Fine-Tuned Chat Models. *ArXiv, abs/2307.09288*.

Dai, Z.; Yang, Z.; Yang, Y.; Carbonell, J.; Le, Q.; and Salakhutdinov, R. 2019. Transformer-XL: Attentive language models beyond a fixed-length context, Proceedings of the Annual Meetings of the Association for Computational Linguistics.

Wang, S.; Li, B.Z.; Khabsa, M.; Fang, H.; and Ma, H. 2020. Linformer: Self-Attention with Linear Complexity. *ArXiv, abs/2006.04768*.

Beltagy, I.; Peters, M.E.; and Cohan, A. 2020. Longformer: The Long-Document Transformer. *ArXiv, abs/2004.05150*.

Kitaev, N.; Kaiser, L; and Levskaya, K. 2020. Reformer: The efficient transformer. *ArXiv, abs/2001.04451*.

Choromanski, K.; Likhosherstov, V.; Dohan, D.; Song, X.; Gane, A.; Sarlos, T.; Hawkins, P.; Davis, J.; Mohiuddin, A.; Kaiser, L.; Belanger, D.; Colwell, L.J.; and Weller, A. 2021. Rethinking attention with performers, ICLR.

Hawthorne, C.; Jaegle, A.; Cangea, C.; Borgeaud, S.; Nash, C.; Malinowski, M.; Dieleman, S.; Vinyals, O.; Botvinick, M.M.; Simon, I.; Sheahan, H.; Zeghidour, N.; Alayrac, J.; Carreira, J.; and Engel, J. 2022. General-purpose, long-context autoregressive modeling with Perceiver AR, ICML.

Gu, A.; Goel, K.; and Ré, C. 2022. Efficiently Modeling Long Sequences with Structured State Spaces, arXiv:2111.00396v3.

Ma, X.; Zhou, C.; Kong, X.; He, J.; Gui, L.; Neubig, G.; May, J.; and Zettlemoyer, L. 2023. Mega: Moving Average Equipped Gated Attention, arXiv:2209.10655v3.

Y. Fu, D.; Dao, T.; Saab, K.; Thomas, A.; Rudra, A.; and Ré, C. 2023. Hungry Hungry Hippos: Towards Language Modeling with State Space Models, arXiv:2212.14052v3.

Zhu, C.; Ping, W.; Xiao, C.; Shoeyb, M.; Goldstein, T.; Anandkumar, A.; and Catanzaro, B. 2021. Long-Short Transformer: Efficient Transformers for Language and Vision, 35th Conference on Neural Information Processing Systems.

Liu, N.F.; Lin, K.; Hewitt, J.; Paranjape, A.; Bevilacqua, M.; Petroni, F.; and Liang, P. 2023. Lost in the Middle: How Language Models Use Long Contexts. *Transactions of the Association for Computational Linguistics, 12*, 157-173.

Singh, S.; and Mahmood, A. 2021. The NLP Cookbook: Modern Recipes for Transformer Based Deep Learning Architectures. *IEEE Access*.

Ji, H.; Zhang, R.; Yang, Z.; Hu, Z.; and Huang, M. 2022. LaMemo: Language Modeling with Look-Ahead Memory. *North American Chapter of the Association for Computational Linguistics*.

Martins, P.H.; Marinho, Z.; and Martins, A.F. 2022. ∞-former: Infinite Memory Transformer-former: Infinite Memory Transformer. *Proceedings of the 60th Annual Meeting of the Association for Computational Linguistics (Volume 1: Long Papers)*.

Gu, A.; and Dao, T. 2023. "Mamba: Linear-Time Sequence Modeling with Selective State Spaces." *ArXiv* abs/2312.00752.

Beck, M.; Poppel, K.; Spanring, M.; Auer, A.; Prudnikova, O.; Kopp, M.K.; Klambauer, G.; Brandstetter, J.; and Hochreiter, S. 2024. xLSTM: Extended Long Short-Term Memory. *ArXiv, abs/2405.04517*.

# Appendix

## A. Further Details on our Enhanced Caching Transformer

In our caching protocol we compress and dynamically retrieve the most relevant compressed segments for any given input. Based on the design constraints an appropriate amount of input sequence compression is performed. Thereafter the sequence is split into the desired segments, and we choose the most similar segments for each query and retrieve them in the original uncompressed form. It ensures only the most relevant information is being picked. This not only helps in reducing the context size, but it also enables in preserving key information. This enhanced caching attention technique is explained in greater detail in the subsequent sections.

### A.1 Enhanced Caching Attention

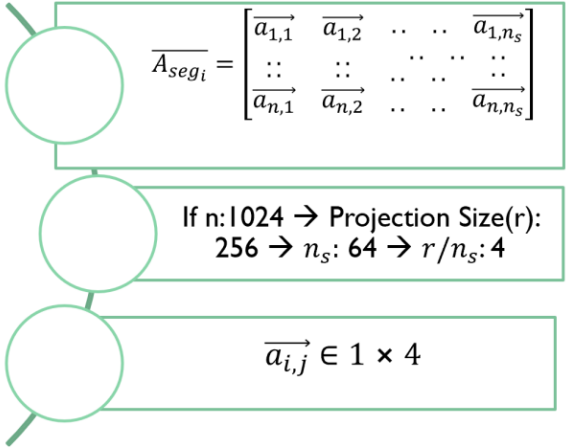

**Figure 5.** Downsized Compression of Attention Matrix along $K_c$, $V_c$

Consider the length of the input sequence to be 1024 tokens that need to be compressed and down projected to 256 tokens. Here we choose to divide the row into ($n_s$) 64 segments. This will yield to a compression ratio ($r/n_s$) of 4. The attention matrix will be of size $\overline{A_{seg_i}} \in \mathbb{R}^{n \times r}$. Therefore for $n_s$ segments, each row in $\overline{A_{seg_i}}$ will consist of row vectors with size $r/n_s$.

Further, the magnitude of the vector $\overrightarrow{a_{i,j}} \in \mathbb{R}^{1 \times r/n_s}$ will represent the attention of the $i^{th}$ word token to the $j^{th}$ compressed segment in the long attention as shown in Figure 5. Thereafter, we compute the root mean square for each of the (1 × 4) sized attention vectors $\overrightarrow{a_{i,j}}$, hence the dimension across each row is downsized from 256 to 64. We use this size for the subsequent attention processing steps as demonstrated in the following section.

### A.2 Averaging in Segment Caching

Attention computation and top-k segment retrieval across all 1024 rows turned out to be computationally cumbersome and time intensive. Therefore, to achieve execution efficiency, we averaged all 1024 input vectors across $p$ consecutive rows for the previous attention matrix $\overline{A_{seg_i}} \in \mathbb{R}^{n \times r}$ where $p$ is a hyperparameter.

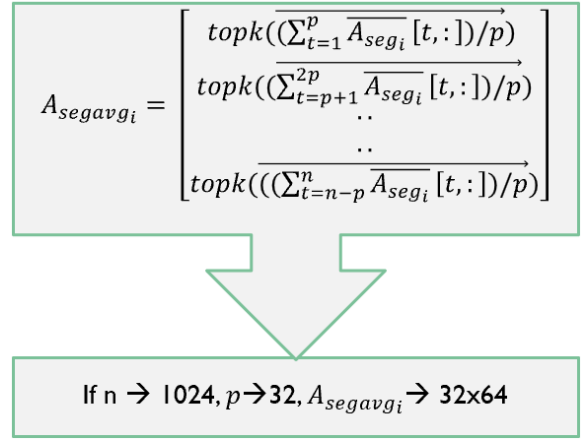

**Figure 6.** Averaged Compression of Attention Matrix along the Input Length

This segment attnetion matrix is further reshaped and compressed into $A_{segavg_i} \in \mathbb{R}^{m \times n_s}$, where $m = n/p=32$ as shown in Figure 6. This implementation was key for our model to achieve superior results outperforming other popular language models of similar size as mentioned in Table 2 and resulted in a faster run time as well.

### A.3 Top-k Retrieval in Segment Caching

Post the compression and averaging, the top $k$ most similar segments were chosen to be retrieved by the order of the attention magnitude betweeen the modified input and key/value matrices. These segments were picked corresponding to each row $m$, which is an averaged input sequence of 32 consecutive words (aveeraged down from 1024) from the segment attention matrix $A_{segavg_i}$.

The hyperparameter $k$ is chosen based on the performance needs and based on that value along with the $k^{th}$ segment, we also extract one segment before and after the $k^{th}$ attentive segment.

Therefore, we define $u$ as the hyperparameter that regulates the number of adjacent segments around $k$ that need to be retrieved from the sequence. For instance, with $k = 5$ and

$u = 3$ will result in a total of 15 uncompressed extracted segments of length 16 from each row as shown in Figure 7.

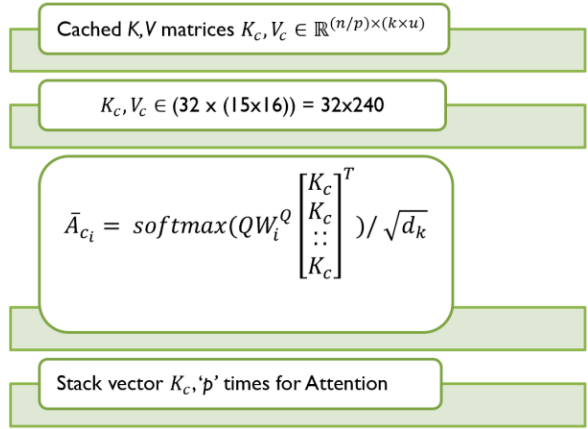

**Figure 7.** Enhanced Attention Matrix after top-k retrieval

**A.4 Overlapping Segments in Long Attention**

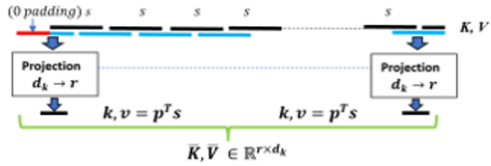

**Figure 8.** Long Attention with Overlapping Segments

As discussed earlier that the segmentation of input into chunks leads to fragmentation of long-term information. This becomes a challenge in building long term dependency. This issue hasn't been addressed in prior Transformer based language models. Therefore, we augment the long attention with segments with a 50% overlap to maintain the continuity of data as shown in Figure 8. The model is trained with the overlapping data as the query that needs to learn the original chunks as key and values.

**A.5 Aggregated Enhanced Long Short Attention**

Thereafter we add the overlapping attention $\bar{A}_{o_i}$ to the long cache attention $\bar{A}_{l_i}$ who have similar shapes. The sliding window (short) attention $\bar{A}_{s_i}$ and our caching attention $\bar{A}_{c_i}$ are concatinated to the above summed attention as pictorially demonstrated in Figure 9.

Here ∥ indicates the catenation of different attentions, $w$ is the window size in short i.e., sliding window attention, $r$ is the projection size in compressing the long attention, $k$ is the $top\ k$ factor in retrieving high attention top $k$ segments, $s$ is the segment size in long attention, $u$ determines the number of segments to be retrieved adjacent to the top $k^{th}$ one.

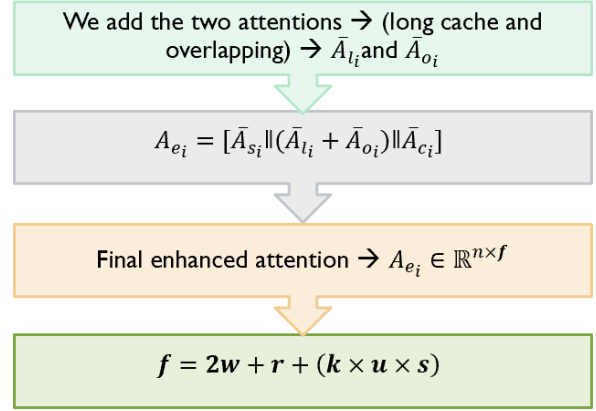

**Figure 9.** Complexity of the Enhanced Attention

Finally, Figure 10 shows the four attention mechanisms that are simultaneously aggregated and successfully inducted in our model architecture.

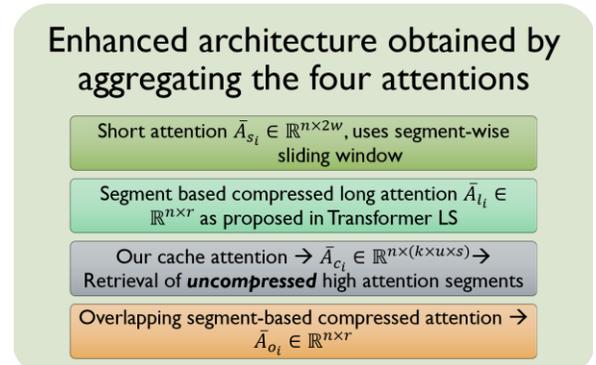

**Figure 10.** Aggregated Enhanced Attention